\documentclass[conference]{IEEEtran}
\IEEEoverridecommandlockouts
\usepackage{cite}
\usepackage{amsmath,amssymb,amsfonts}
\usepackage{algorithmic}
\usepackage{graphicx}
\usepackage{textcomp}
\usepackage{csvsimple}
\usepackage{booktabs}
\usepackage{xcolor}
\usepackage[hyphens]{url} 
\usepackage{listings}
\usepackage{eurosym}
\usepackage{subcaption}

\def\BibTeX{{\rm B\kern-.05em{\sc i\kern-.025em b}\kern-.08em
    T\kern-.1667em\lower.7ex\hbox{E}\kern-.125emX}}

\begin{document}

\title{General Board Game Concepts}

\author{\IEEEauthorblockN{{\'E}ric Piette, Matthew Stephenson, Dennis J.N.J. Soemers and Cameron Browne}
\IEEEauthorblockA{\textit{Department of Data Science and Knowledge Engineering} \\
\textit{Maastricht University}\\
Maastricht, the Netherlands \\
\texttt{\{eric.piette,matthew.stephenson,dennis.soemers,cameron.browne\}@maastrichtuniversity.nl}}
}

\maketitle

\begin{abstract}
Many games often share common ideas or aspects between them, such as their rules, controls, or playing area. 
However, in the context of General Game Playing (GGP) for board games, this area remains under-explored. We propose to formalise the notion of ``game concept", inspired by terms generally used by game players and designers. Through the Ludii General Game System, we describe concepts for several levels of abstraction, such as the game itself, the moves played, or the states reached. This new GGP feature associated with the ludeme representation of games opens many new lines of research. The creation of a hyper-agent selector, the transfer of AI learning between games, or explaining AI techniques using game terms, can all be facilitated by the use of game concepts. Other applications which can benefit from game concepts are also discussed, such as the generation of plausible reconstructed rules for incomplete ancient games, or the implementation of a board game recommender system.

\end{abstract}


\section{Introduction}

Since the foundation of Artificial Intelligence (AI), games have often been used as testbeds for major advancements \cite{gameaibook}. This can be explained by their simplicity and popularity, but also because, as humans, we can perceive the problems (game interfaces) and their solutions (strategies). 
In the field of game AI, two agent types can be distinguished. Dedicated Game AI agents designed for a single game (or a small set of games) and General Game AI agents built to play any arbitrary game. 

Game-specific agents have been studied since the beginning of the AI field \cite{Samuel59,Samuel60} and are often used as benchmarks of well-known problems. For example, many single-player games, commonly called puzzles, are interesting illustrations of planning problems (e.g. Rubik's Cube \cite{Korf97} or Sokoban \cite{Junghanns01}) or Constraint Satisfaction Problems \cite{PietteCog19}. Similarly, multi-player games are representations of many different AI research areas (Knowledge Representation, Reinforcement Learning, Transfer Learning, etc.) which lead to the development of many game AI agents trying to outperform human players. 
Nowadays, computer players for classical board games like Chess \cite{stockfish17}, Go \cite{silver16}, Shogi \cite{shogiSilver} or Checkers \cite{checkers}, can defeat expert human players. Even for stochastic games like Backgammon \cite{tdgammon}, games with incomplete information like Poker \cite{pokerreview}, or video games like Starcraft II \cite{starcraft}, computer players have reached an excellent level of play.
The success of these dedicated game-playing programs has demonstrated their capacity to outperform humans, marking a milestone in artificial intelligence research \cite{deepblue, silver16}. However, they are often highly reliant on game-specific knowledge and expertise, preventing them from playing other games as effectively. 

Building general game AI agents that perform well on any board game, rather than being specialised, is the aim of General Game Playing (GGP). These general game AI agents are implemented without knowing the games they will play in advance and consequently, they cannot use specific pre-defined knowledge about them when playing the game. This differs greatly from dedicated game AI agents, as the AIs developed for specific games are often biased by human-developed heuristics. 
Unfortunately, humans still tend to outperform AI in general board games.
This can be explained by the human capacity for understanding common aspects between games and consequently being able to learn new games quicker by detecting common points with previously played games. 

This paper introduces board game concepts for GGP through the Ludii general game system. These concepts are expressed in game terms commonly used by game players and designers, making them an interesting mechanism for providing human-understandable explanations of different AI techniques. These concepts can be associated with several levels of abstraction, such as the game itself, a game trial, a game state, or individual moves. This opens many new possibilities for different AI research topics, including agent selection, transfer learning between games, AI explainability, as well as game generation, reconstruction, and recommendation.

The remainder of this paper is organised as follows: Section \ref{Background} describes the field of GGP and the different general game systems proposed so far. Section \ref{Ludii} describes the Ludii system as well as the ludeme-based language used to describe the games and the reasons behind its use for identifying game concepts. Section \ref{Concepts} highlights the notion of game concepts in the context of Ludii, how they are modelled and computed for each game. Section \ref{ResearchDirections} proposes several GGP research directions which can benefit from our proposed game concepts. Finally, Section \ref{Conclusion} concludes the paper.

\section{Background}
\label{Background}

\subsection{General Game Playing}
\label{GGP}

In Artificial Intelligence, the General Game Playing (GGP) challenge \cite{Genesereth2014} is to develop computer players that understand the rules of previously unknown games, and learn to play these games well without human intervention in situations where its knowledge of the games rules, objectives and strategies is limited to the game description. Developing these general agents, and the AI techniques behind them, is vitally important in the development of real-world agents which can deal with unpredictable and novel situations. For this reason, General Game Playing is seen as a necessary step on the way to Artificial General Intelligence (AGI) \cite{ArtificialGeneralIntelligence}.

Historically, the first GGP model is defined in 1968 by Jacques Pitrat \cite{Pitrat68} to describe two-player games with complete information and rectangular boards. Then, only from the '90s, other work on General Game systems appeared such as SAL \cite{SAL}, Metagamer \cite{METAGAMER}, Hoyle \cite{HOYLE}, and Morph-II \cite{MorphII}. After the turn of the millennium, more modern General Game systems started appearing such as Multigame \cite{Multigame} in 2001 and Zillions of Games \cite{zillions} in 2002. 

\subsection{General board Game Languages}
\label{GGPSystems}

Many GGP systems describe their games using a standardised game description language. Different game descriptions are written in different formats and with different levels of abstraction, lending themselves to different kinds of knowledge representation, reasoning, and learning approaches (such as for performance reasons).

\subsubsection{GDL}

Since 2005, the \textit{Game Description Language} (GDL) \cite{genesereth05} and the GGP-Base platform \cite{ggpbase} proposed by the Stanford Logic Group\footnote{Stanford Logic Group: \url{logic.stanford.edu/}} has become one of the main standards for academic research in GGP. The purpose of GDL is to provide a generic language for representing any board game \cite{thielscher2011}, including collaborative games and games with simultaneous actions. GDL describes the games in a language inspired by Prolog, using first-order logical clauses. An extension named GDL-II \cite{schiffel14} has been developed to handle games with partial observations and stochastic actions, and another extension named GDL-III \cite{thielscher17} for epistemic games.
This formalism and platform have provided a high-level challenge which has led to important research contributions \cite{swiechowski15,bjornsson16} -- especially in {\it Monte Carlo tree search} (MCTS) enhancements \cite{finnsson08,finnsson10}, with some original algorithms combining constraint programming, MCTS, and symmetry detection \cite{koriche17}. These techniques have lead to the development of several General Game AI agents that perform well on specific games (e.g. Gamer \cite{Gamer}, CadiaPlayer \cite{Cadiaplayer}, ClunePlayer \cite{Clune}, WoodStock \cite{WoodStock}, etc.).



Unfortunately, GDL also has some negative points. This includes poor efficiency \cite{Piette2020Ludii}, the fact that each element of the game has to be defined tabula rasa and cannot easily be taken from previously modelled examples, and verbose game descriptions that are not representative of any game-related aspects that human players typically use. 
The opaque and low-level character of GDL descriptions, without any shared semantics or high-level concepts being explicitly recognisable between games, forms a challenge for tasks such as the creation of mappings between games for transfer learning, or the generation of human-understandable explanations about any aspect of a game.

\subsubsection{RBG}

The Regular Boardgames (RBG) system \cite{Kowalski19} proposes the idea of encoding piece movement for games using a regular language. This formalism can describe any finite deterministic game with perfect information excluding simultaneous moves. RBG game descriptions are typically much short than their GDL counterparts, making it much easier to model complex board games (e.g. Chess, Go and Arimaa). RBG also runs substantially faster than GDL, averaging over 50 times as many playouts during a recent comparison \cite{Piette2020Ludii}.

\subsubsection{Ludii}

The Ludii system\footnote{The source code of Ludii is available at \url{github.com/Ludeme/Ludii}}, named after its predecessor {\sc LUDI} \cite{browne09}, is a complete general board game system, that can model games as a tree of ludemes. The decomposition of games into their conceptual units of game-related information {\it ludemes} \cite{parlett16} results in more intuitive games descriptions compared to prior GGP systems. The number and variety of board games that are implemented within Ludii also go far beyond those presented by most prior general game systems \cite{LudiiCompetition}. Ludii is capable of modelling the full range of playable GDL games, as well as stacking games, boardless games, and games with hidden information. This, combined with high efficiency compared to other general systems \cite{Piette2020Ludii}, makes Ludii the ideal system for investigating the new game concepts-based research directions discussed in this paper.

\subsection{Related Work}
\label{RelatedWorks}

In the context of game AI research, prior work on feature extraction has been predominantly focused on video games. This includes more general frameworks such as the General Video Game AI (GVGAI) system \cite{gvgaioverview, Bontrager2016}, but also specific video games such as Angry Birds \cite{Stephenson17} and Starcraft \cite{6633643}. These papers used their extracted features to predict the performance of different agents on unknown games, creating a portfolio or ensemble agent that combines the strengths of multiple AI techniques.

Despite the demonstrated advantages of this approach, little work has been done on feature extraction for general board games. Banerjee and Stone \cite{ValueFunctionTransferGGP2} proposed a reinforcement learning game player interacting with the GGP-base system, which can transfer knowledge learned in one game to expedite learning in other games. This work is based on value-function transfer \cite{ValueFunctionTransferGGP} where general features are extracted from the state space of a previous game and matched with the different state space of a new game. This work showed good performance for low-level features on small games, but only for a limited set of specific types of source-target pairs where appropriate state space mappings can be automatically detected based on GDL game descriptions \cite{Kuhlmann2007GraphBased}.

Some of the best General Game Playing players based on GDL have shown the importance of heuristics and game features by using them to improve their performance during GGP competitions. ClunePlayer used heuristic evaluation functions to represent simplified games as abstract models, incorporating the most essential aspects of the original game to construct tailored heuristics \cite{Clune}. CadiaPlayer used template matching to identify simple board game features, such as square tiling or specific piece types \cite{Cadiaplayer}. Each of these agents won the International General Game Competition \cite{genesereth13}, in 2005 for ClunePlayer and 2007, 2008, and 2012 for CadiaPlayer.

\section{Ludii}
\label{Ludii}

Ludii\footnote{Ludii is available at \url{ludii.games/download.php}} uses a {\it class grammar} approach~\cite{CameronB16} to provide a direct link between the keywords in its game descriptions and the underlying Java code that implements them. The core of Ludii is a ludeme library, consisting of several classes, each implementing a specific ludeme. Ludemes are used to define both the {\it form} of the game (rules and equipment) and its {\it function} (legal moves and outcomes for the end state).

Each game description is composed of three different sections which define the relevant information about the players (number, facing direction, etc.), the equipment (board, pieces, etc.), and the rules of the game. The rules of the game can also be decomposed into four specific sub-sections:
\begin{itemize}
    \item \textbf{Meta-rules} applied to each state (e.g. no repetition).
    \item \textbf{Starting rules} defining the initial state.
    \item \textbf{Playing rules} defining the legal moves for a state $s$.
    \item \textbf{Ending rules} defining the conditions under which the game's outcome is determined.
\end{itemize}

  \begin{figure*}[!t]
\centering
    \begin{subfigure}{0.33\textwidth}
    \includegraphics[width=1.05\textwidth]{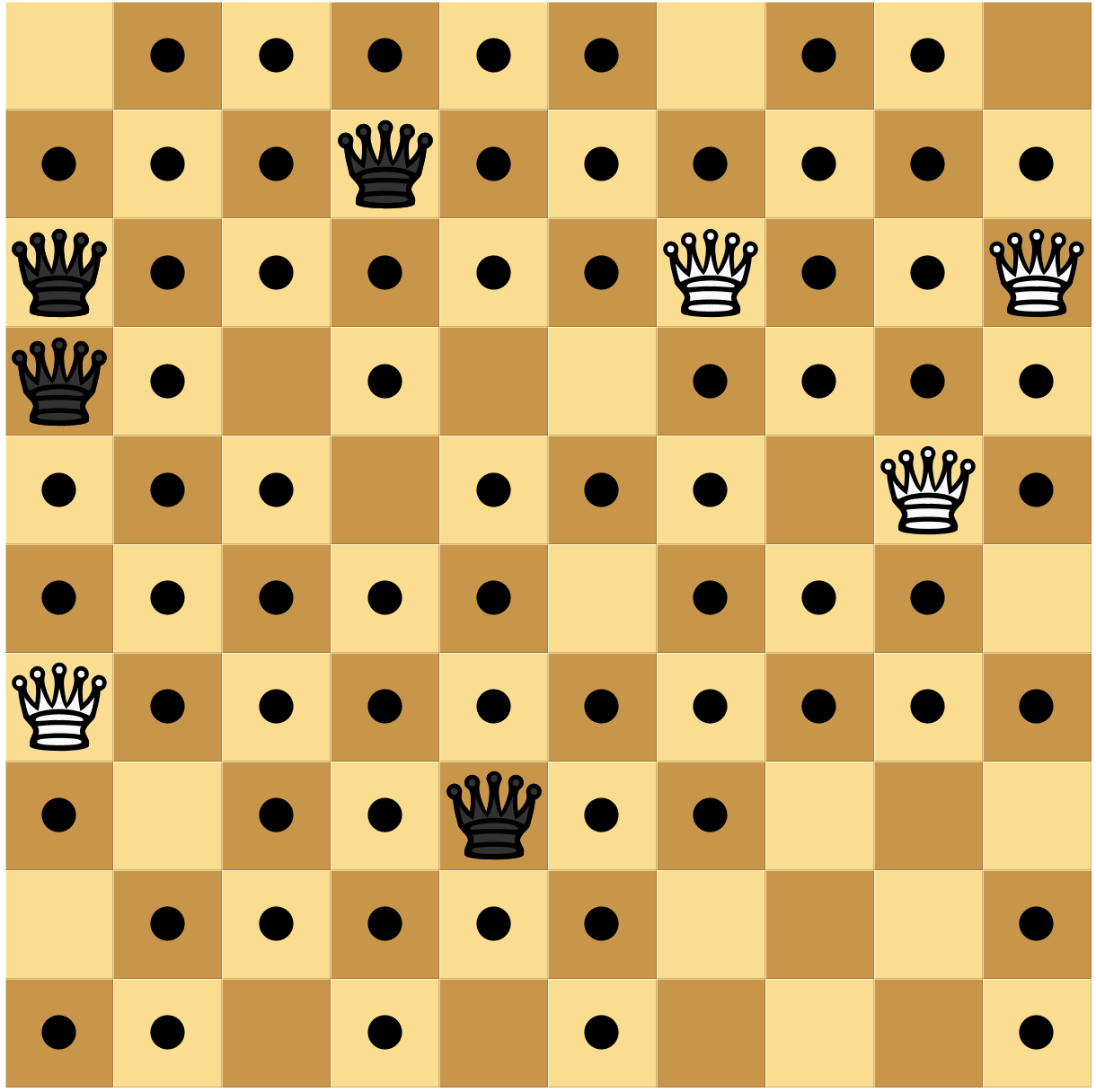}
  \end{subfigure}
      \begin{subfigure}{0.05\textwidth}
    ~
  \end{subfigure}
  \begin{subfigure}{0.58\textwidth}

\footnotesize
\lstset{basicstyle=\ttfamily}
\begin{lstlisting}
(game "Amazons"  
    (players 2)  
    (equipment { 
        (board (square 10)) 
        (piece "Queen" Each (move Slide (then (moveAgain))))
        (piece "Dot" Neutral)
    })  
    (rules 
        (start { 
            (place "Queen1" {"A4" "D1" "G1" "J4"})
            (place "Queen2" {"A7" "D10" "G10" "J7"})
        })
        (play 
            (if (is Even (count Moves))
                (forEach Piece)
                (move Shoot (piece "Dot0"))
            )
        )
        (end (if (no Moves Next) (result Mover Win)))  
    )
)
\end{lstlisting}
  \end{subfigure}

        
    \begin{subfigure}{0.35\textwidth}
    \includegraphics[width=1.1\textwidth]{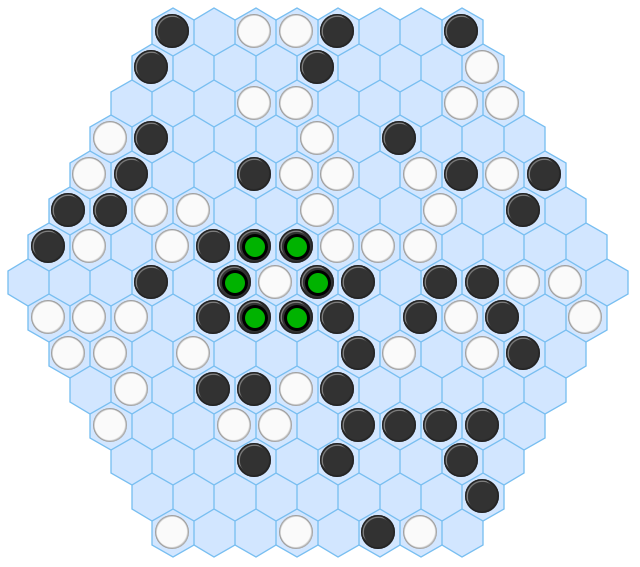}
  \end{subfigure}
      \begin{subfigure}{0.05\textwidth}
    ~
  \end{subfigure}
  \begin{subfigure}{0.58\textwidth}

\footnotesize
\lstset{basicstyle=\ttfamily}
\begin{lstlisting}
(game "Havannah"  
    (players 2)  
    (equipment { 
        (board (hex 8)) 
        (piece "Marker" Each) 
    })  
    (rules 
        (play (move Add (to (sites Empty))))
        (end 
            (if 
                (or {
                    (is Loop)  
                    (is Connected 3 SidesNoCorners) 
                    (is Connected 2 Corners) 
                })
                (result Mover Win)
            )
        )    
    )
)
\end{lstlisting}
  \end{subfigure}
  
  \caption{Completed games of Amazons and Havannah, on the Ludii system, along with their ludeme-based game descriptions.}
  \label{Fig:LudiiGames}
\end{figure*}

An important benefit of this representation is that it allows each game description to be expressed in a compact and human-understandable manner by hiding unnecessary implementation details \cite{logicGuide}. 
Figure \ref{Fig:LudiiGames} shows Ludii game descriptions for the board games Amazons\footnote{Game of the Amazons: \url{en.wikipedia.org/wiki/Game_of_the_Amazons}} and Havannah,\footnote{Havannah: \url{en.wikipedia.org/wiki/Havannah}} alongside the graphical interfaces of playouts run on these games.

Moves in Ludii are represented using an atomic model. At each state $s$, the ludemes describing the playing rules are evaluated and return a list of $k$ legal moves $\mathcal{M}$: $\langle m_1, \ldots, m_{i}, \ldots, m_k \rangle$. In Ludii, the transition between two successive states $s_i$ and $s_{i+1}$ is possible thanks to a sequence of atomic actions $A_i$ applied on $s_i$. Such a sequence is modelled as a move $m$: $\langle a_1, \ldots, a_{i}, \ldots, a_n \rangle$ with $n$ the number of actions in $A_i$.
Thanks to this representation, it is possible to associate with each move $m$ the concepts triggered by the list of actions defining $m$. Similarly, concepts can be associated with any state $s$, by computing the concepts involved at $s$.

\section{Game Concepts}
\label{Concepts}

A concept is an abstract representation that can be shared between different objects or ideas highlighting common points. Previous work (see Section \ref{RelatedWorks})) commonly uses the term ``game feature'' to designate a high-level aspect of games. However, in a board game context but also for the more general purpose of this paper, we are going to use the term ``game concept'' due to the existing use of the term ``feature'' by previous work on state-action features for board games \cite{featureKEG}.

Every game concept in Ludii is defined by its name, its category, its data type (numerical or binary) and its computation type. Each concept's name is selected to be as close as possible to terms used by human players or game designers. Version 1.2.0 of Ludii, used in this paper, currently implements 428 distinct game concepts. 
To organise these concepts, we propose a taxonomy\footnote{The proposed taxonomy can be found at \url{ludii.games/searchConcepts.php}} inspired by the Core Subject Taxonomy for Mathematical Sciences Education outlined by the Mathematical Association of America \cite{maa2004}. 

\subsection{Categories}

The seven main concept categories are:
\begin{itemize}
    \item \textbf{Properties}: Concepts related to the format of the game (time model, information type, symmetries, players, \ldots). 
    \item \textbf{Equipment}: Concepts related to the board (shape, tiling, graph, \ldots) and pieces (tile, dice, large piece, \ldots).
    \item \textbf{Rules}: Concepts related to each rule type (meta, start, play, end).
    \item \textbf{Math}: Concepts relating to fields of Mathematics (Arithmetic, Comparison, Logic, Algorithmic, \ldots).
    \item \textbf{Metrics}: Concepts describing well-know game metrics (game length, branching factor, \ldots).
    \item \textbf{Visual}: Concepts describing a game's graphical style.
    \item \textbf{Implementation}: Concepts describing game implementation details.
\end{itemize}
Thanks to this taxonomy, it is possible to focus only on a specific subset of categories depending on the specific application or research field. For example, many game-playing applications would have no use for the visual category, while explainable AI research will likely disregard the implementation category.

\subsection{Data Type}

A game concept can be numerical or binary. Binary concepts are used to show the existence of concepts in games, while numerical concepts are used to quantify them.

A binary game concept is activated if a specific ludeme, or a combination of ludemes, is used in the game's description. As an example, the concept {\sc Stochastic} describes games involving chance elements. It can be activated by multiple ludemes in isolation, such as rolling dice {\sc (roll)} or the use of any random value {\sc (value Random 1 5)}. Contrary to this, the concept {\sc Hop Capture} will be triggered if the game involves a {\sc (move Hop ...)} ludeme and a capturing effect within it, such as the ludeme {\sc (remove ...)}. 

Rather than being activated or not, numerical game concepts are instead associated with a value. This can be an integer, such as the number of players ({\sc Num Players}) or the number of playable sites ({\sc Num Playable Sites}), or a float, such as the average number of possible directions per site ({\sc Num Directions}). Some numerical concepts also correspond to the frequency of binary concepts, such as the average number of times a specific terminal state is reached, e.g., in Havannah, the frequency of winning with a loop ({\sc Loop End}) compare to connected regions ({\sc Connection End}), or the average number of times a specific move type was made, e.g. In Amazons, the frequencies of the concepts {\sc Slide} or {\sc Shoot}.

\subsection{Computation}

Within the Ludii software, game concepts can be obtained in two different ways. One is done near-instantaneously during game compilation (compilation concepts), the other requires running playouts (playout concepts). This distinction is important depending on the intended application. For example, tasks that have to be responsive cannot use the concepts that require playouts, as these take significantly longer to compute.

Compilation concepts are based only on the ludemes used to describe static properties of the game, such as the dimensions of the game board. During the compilation process of Ludii games \cite{CameronB16}, the existence of specific ludemes or combinations of ludemes trigger each of these concepts. Binary concepts are simply activated by the existence of ludemes while numerical concepts instead have their values set, such as the number of component types ({\sc Num Component Types}).

Each atomic action in Ludii can return the concepts triggered by their application to a game state, consequently, game concepts can be associated with each move played instantaneously, due to the atomic-action representation of a move. However, at a higher level such as the game, playout concepts require more time to be computed. All concepts relating to the frequency of specific binary concepts are based on playouts. Each playout corresponds to a sequence of moves called a trial $\tau$: $\langle m_1, \ldots, m_{i}, \ldots, m_{ter} \rangle$ with $m_{ter}$ the last move played reaching the terminal state $s_{ter}$. Thanks to the association of each move to a set of binary concepts, each trial $\tau$ can compute the frequency of the concepts involved. By running many playouts and consequently obtaining many trials for a single game, the values of the frequency game concepts can be computed by averaging the frequency of the concepts in each trial. Similarly, all the concepts belonging to the Metrics category, such as {\sc Game Length} or {\sc Branching Factor}, are also computed using these playouts. For all playout concepts, it is important to note that their values are dependent on the playout type, which can be obtained randomly or with different game-playing agents.

 \begin{table*}[t]
 \caption{Selection of game concepts detected on some popular games.}
 \label{Table:ConceptsOnGames}
 \begin{center}
 \begin{tabular}{@{}lccccccccc@{}}
 \toprule
 \textbf{Game} & \textbf{Players} & \textbf{PlayableSites} & \textbf{Checkmate} & \textbf{SquareTiling} & \textbf{HexTiling} & \textbf{AddMove} & \textbf{Capture} & \textbf{ConnectionEnd} & \textbf{Stochastic}  \\
 \midrule
 Amazons            & 2 & ~64    & $\times$    & \checkmark & $\times$      & \checkmark   & $\times$    & $\times$     & $\times$ \\
 Backgammon         & 2 & ~28    & $\times$    & $\times$   & $\times$      & $\times$     & \checkmark  & $\times$     & \checkmark \\
 Chess              & 2 & ~64    & \checkmark  & \checkmark & $\times$      & $\times$     & \checkmark  & $\times$     & $\times$ \\
 Chinese Checkers   & 6 & 121   & $\times$    & $\times$   & \checkmark    & $\times$     & $\times$    & $\times$     & $\times$ \\
 Go (19x19)         & 2 & 361   & $\times$    & \checkmark & $\times$      & \checkmark   & \checkmark  & $\times$     & $\times$ \\
 Havannah           & 2 & 169   & $\times$    & $\times$   & \checkmark    & \checkmark   & $\times$    & \checkmark   & $\times$ \\
 Hex (11x11)        & 2 & 121   & $\times$    & $\times$   & \checkmark    & \checkmark   & $\times$    & \checkmark   & $\times$ \\
 Oware              & 2 & ~14    & $\times$    & $\times$   & $\times$      & $\times$     & \checkmark  & $\times$     & $\times$ \\
 Shogi              & 2 & ~95    & \checkmark  & \checkmark & $\times$      & $\times$     & \checkmark  & $\times$     & $\times$ \\
 Snakes and Ladders & 4 & 100   & $\times$    & \checkmark & $\times$      & $\times$     & $\times$    & $\times$     & \checkmark \\
 Tic-Tac-Toe        & 2 & ~~9     & $\times$    & \checkmark & $\times$      & \checkmark   & $\times$    & $\times$     & $\times$ \\
 Xiangqi            & 2 & ~90    & \checkmark  & \checkmark & $\times$      & $\times$     & \checkmark  & $\times$     & $\times$  \\
 \bottomrule
 \end{tabular}
 \end{center}
 \end{table*}

 Table \ref{Table:ConceptsOnGames} shows the values for a subset of concepts on several popular games. Due to the large number of possible game concepts, we refer readers to the webpage of each available Ludii game at \url{ludii.games/library.php}. Each game entry lists all the compilation concepts that were detected, while playout concepts were computed by running 10,000 random playouts for each game. For example, for Amazons and Havannah, the games shown in Figure \ref{Fig:LudiiGames}, the concepts are shown at \url{ludii.games/concepts.php?gameId=52} and \url{ludii.games/concepts.php?gameId=372}, respectively. We can observe that for Amazons the frequencies of the concepts {\sc Slide} and {\sc Shoot} are both $50\%$, while for Havannah the frequencies of the concepts {\sc ConnectionEnd} and {\sc LoopEnd} are $27\%$ and $73\%$, respectively.


\section{GGP Research Directions}
\label{ResearchDirections}

Having now described the different game concept categories and data types, as well as how they are computed, we turn to the different possible applications and research directions that can utilise them.

\subsection{Agent Selection}

Hyper-heuristic approaches aim to select the best agent, heuristic or strategy for a given task from a pre-defined set of choices \cite{hyperSurvey}. Such approaches, especially for game AI research, typically involve identifying features between games that give an indication of how these different choices will perform \cite{Mendes2016HyperheuristicGV,7017583}. Our proposed game concepts may provide such features, if a correlation between the values for certain concepts and the performance of different agents can be identified.

Although not yet proven, it is highly likely that the existence of certain concepts or concept values would affect the abilities of different game-playing agents. A larger board and more pieces would typically indicate a greater branching factor, which affects the performance of certain AI techniques more than others. Likewise, agents that require full playouts to compare different moves (e.g. Monte Carlo tree search) may perform worse on games that take longer to complete, compared to agents that use defined state evaluation heuristics (e.g. Minimax).

After determining the performance of a set of agents for a given set of games, we can attempt to train models to predict the best agent for a game based on its concepts. If successful, this can provide a portfolio agent which can predict the best agent on a new game using its concept values. A similar approach has been previously presented in \cite{stephenson2021general}, which uses the ludemes within the games -- rather than their concepts -- to predict the performance of different general-game-playing heuristics. It is likely that the game concepts provide additional information which can help give more accurate predictions, thus leading to better portfolio agent performance.

\subsection{Transfer Learning}

Transfer learning research in games, and more generally reinforcement learning \cite{Taylor2009TransferRL,Lazaric2012TransferRL,Zhu2020TransferDeepRL}, focuses on transferring learned objects such as heuristics, value functions, policies, etc. from one or more source domains, to one or more target domains. This may involve transferring trained weights and/or representations -- which may be explicit features \cite{featureKEG}, or learned representations such as those in hidden layers of deep neural networks.

Transfer of heuristics or value functions, which are functions of game states, between any pair of games requires that those games have identical state spaces, or compatible state representations such that we can create mappings between their state spaces. Similarly, transfer of policies, which are functions of game states and actions, also requires identical action spaces or compatible action representations that enable mappings between action spaces. There has been some research towards transferring value functions in GGP based on general features of the shape of a lookahead search tree \cite{ValueFunctionTransferGGP2}, as well as value function transfer in a limited set of specific types of source-target pairs where appropriate state space mappings can be automatically detected based on GDL game descriptions \cite{Kuhlmann2007GraphBased}.

Ludii's object-oriented, game-independent state and action representations \cite{logicGuide} has already been demonstrated to facilitate straightforward mappings between state and action spaces of different games, allowing for effective transfer of deep neural networks with both policy and value heads between many different pairs of games \cite{Soemers2021TransferFullyConv}. However, even when such mappings are possible, transfer learning is sometimes still ineffective (or even detrimental, a phenomenon known as negative transfer \cite{Zhang2020NegativeTransfer}) due to significant differences in the goals or optimal strategies between games. For example, the state and action representations of \textit{Hex} and \textit{Mis{\`e}re Hex} -- which is \textit{Hex} with an inverted win condition -- are very similar, but neither policy nor value functions work well when transferred directly (although learned features, or hidden representations, can still be useful).

Game concepts may be viewed as ``summarising'' a problem or task description for a game, and hence also used to analyse in which aspects any pair of games is similar or different. This can be used to predict whether or not transfer may be successful, or more specifically to predict which aspects -- only policies, or only value functions, or only features/learned representations, etc. -- may successfully transfer \cite{BouAmmar2013AutomatedTransfer,BouAmmar2014MDPSimilarity}.

\subsection{Explainability}

Explainable AI (XAI) is an emerging field in Artificial Intelligence that has become increasingly important in recent years \cite{XAIsurvey}. XAI attempts to bring transparency to how AI agents perform, by providing human-understandable explanations for the decisions they make. 
For many AI agents, the strategies played are generally obscure and difficult to decipher. For example, Deep Reinforcement Learning based agents such as AlphaGo \cite{alphago} outperforming the best Go players, are playing strategies that are still analysed by Go experts \cite{Zhou18} years after its victory on Lee Sedol \cite{Move37}. 
It can be similarly difficult to understand the strategies followed by agents that do not explicitly use understandable domain knowledge, such as general game agents or puzzle solvers.



Correlations between concepts associated with played moves, and concepts associated with the states in which moves are played, may provide human-understandable insights -- expressed in ``game terms'' -- of agents' strategies. For example, identifying specific concepts triggered by moves played only from states always corresponding to the same set of concepts can explain the strategy followed by an AI agent.

For General Game agents, the good or bad performance of AI techniques/agents on games could be explained by the concepts shared between games. This explanation could be provided at a game agent level but also for the different parameters or guiding search techniques to some subsets of games but not others. For example, agents based on MCTS techniques or an Alpha-Beta search without any pre-defined knowledge, are generally achieving different performances based on the rules of the games requiring more or less time to be computing at each state (e.g. Chess-like games are generally in favour of Alpha-Beta agents because of the checkmate rule).

\subsection{Move Evaluation}

Concepts associated with moves can be used as a means of classifying them based on high-level categories, such as distinguishing capture moves from moves that do not capture. This may provide basic evaluations of the usefulness of different moves, and hence used to guide search algorithms. For example, such concepts could be used for move ordering in minimax-based agents, or to bias move selection in playouts using techniques such as FAST \cite{finnsson10}. Such guidance will typically be fairly basic due to the high-level nature of concepts, but also highly efficient because the concepts can be extracted directly from moves generated by Ludii. This process does not require any additional computation, as would be required for features of a successor state.

\subsection{Game Generation}

Board game generation through the use of an evolutionary algorithm has been shown to provide interesting new games \cite{browne09, browne11}. One improvement to this work might be to generate games that result in some particular combination of game concepts that a specific player might find appealing. Game concepts could also be an important aspect of fitness functions for evaluating generated games, particularly for search-based approaches in the same vein as previous work on puzzle generation \cite{puzzleGeneration}. Concepts can also help to direct the selection of specific ludemes within a generated games description, by favouring ludemes that are associated with desirable concepts.



The use of game concepts to generate games can also provide new challenges. For example, the generation of games that use some novel combination of concepts that have not been previously created, perhaps leading to new and exciting gameplay properties. Similarly, the simplification of games by removing a particular concept from its rules could provide games that help teach specific rules to beginners. It has been shown in prior work \cite{Kuhlmann2010} that agents learn faster when independently trained on separate simpler versions of a large game. This is useful in cases where training time is limited, for example in AI competitions \cite{LudiiCompetition, genesereth13}.

\subsection{Game Reconstruction}

As described by the Digital Ludeme Project (DLP) \cite{Browne18ModernTechniquesForAncientGames}, there are a large number of historical games for which the rules are unknown or incomplete. A process similar to that of game generation could be used to reconstruct the rules of an incomplete game. By looking at the concepts used in known nearby games, we can use these to predict which concepts our incomplete game might have. In essence, for each game concept, we can calculate a probability that it was included in an incomplete game, based on the concepts present in other similar games.

\subsection{Game Recommender Systems}

Concepts can also be used to recommend new games to human players based on their preferences. If a player favours games with certain concepts, such as asymmetric games with hidden information, but dislike others, such as games with capture moves, then we can find and suggest other games in Ludii that also contain these desired combinations. Our concepts could also be combined with other board game recommendation systems, such as those developed for the online board game geek database \cite{sym12020210,10.1145/3341105.3375780,10.1145/3308560.3316457,8885455}, to create an even more extensive system.

Game concepts could also be used as searching criteria to find appropriate benchmark games for AI agents. Many AI techniques are not suitable or applicable to certain games types, such as algorithms dedicated to deterministic games versus agents dedicated to stochastic games with hidden information. This applicability problem can be more specific, mainly when the techniques are based on some specific heuristics or functions such as knowledge designed for some specific game goals. The search for good benchmarks can be time-consuming and would be much easier with the use of game concepts showing the properties of games in a recommender system.
Identifying smaller subsets of games that give near equal information about an agent compared to the full set has been previously investigated for the GVGAI framework \cite{9185834}. By comparing the concepts present within the Ludii game corpus, we can achieve a similar means of creating a smaller game subset that still provides maximal concept coverage.



\section{Conclusion}
\label{Conclusion}

This paper refines the notion of game concepts to General Board Game Playing. Such concepts are defined using common terms which are understood by game players and designers. These concepts can be identified automatically for games in Ludii thanks to the ludeme representation, and are associated with any level of game abstraction. 
Game concepts provide several interesting research possibilities, including the creation of a portfolio agent via agent or heuristic selection, transfer learning between games, explainability of AI techniques and strategies, Game Generation, Game Reconstruction and Game Recommendation.

Future work over the next few years will primarily involve identifying more game concepts and implementing their detection within the Ludii system, as well as investigating the research directions proposed in this paper. In the long term, the ludeme philosophy used to represent games could be extended to other domains, such as protein folding, physics simulations or chemical reactions, by adapting the notion of ludemes to conceptual units of each of these topics. Consequently, concepts specific to these fields could be identified and detected by similar techniques, providing a human understandable explanation for any technique applied in these fields.

\section*{Acknowledgment}

This research is part of the European Research Council-funded Digital Ludeme Project (ERC Consolidator Grant \#771292) run by Cameron Browne at Maastricht University's Department of Data Science and Knowledge Engineering. 

\bibliographystyle{IEEEtran}
\bibliography{References}

\end{document}